# Web Services: A Process Algebra Approach

Andrea Ferrara

DIS - Università di Roma "La Sapienza"
Via Salaria 113, 00198 Roma, Italia
Emails: ferrara@dis.uniroma1.it

**Abstract.** It is now well-admitted that formal methods are helpful for many issues raised in the Web service area. In this paper we present a framework for the design and verification of WSs using process algebras and their tools. We define a two-way mapping between abstract specifications written using these calculi and executable Web services written in BPEL4WS. Several choices are available: design and correct errors in BPEL4WS, using process algebra verification tools, or design and correct in process algebra and automatically obtaining the corresponding BPEL4WS code. The approaches can be combined. Process algebra are not useful only for temporal logic verification: we remark the use of simulation/bisimulation both for verification and for the hierarchical refinement design method. It is worth noting that our approach allows the use of any process algebra depending on the needs of the user at different levels (expressiveness, existence of reasoning tools, user expertise).

## 1 Introduction

Web services (WSs) are distributed and independent pieces of code solving specific tasks which communicate with each other through the exchange of messages. A more unusual specificity that distinguishes them from more traditional software components is that they are deployed and then accessed through the internet. Some XML-based standardized technologies have already been proposed to support WSs development: WSDL interfaces abstractly describe messages to be exchanged, SOAP is a protocol for exchanging structured information, UDDI is used to publish and discover WSs, BPEL4WS (BPEL for short) is a notation for describing executable business process behaviors. WSs raise many theoretical and practical issues which are part of on-going research. Some well-known problems related to WSs are to specify them in an adequate, formally defined and expressive enough language, to compose them (automatically), to discover them through the web, to ensure their correctness.

Formal methods provide an adequate framework (many specification languages and reasoning tools) to address most of these issues (description, composition, correctness).
Different proposals have emerged recently to abstractly describe WSs, most of which are grounded on transition system models (Labelled Transition Systems, Mealy automata, Petri nets, etc.) [5,14,23,12,19,11].

However, very few approaches have been proposed to help the design and then development of WSs, especially from these abstract descriptions (as done in the classical software engineering lifecycle to develop software systems). As one of the most related example, existing works [23,10,22,11] aimed at giving abstract representations to e-services in order to ensure properties on a bundle of interacting services. Respect these works, we use process algebras (PAs for short) as abstract representation. Process algebras offer more respect to all these previous approaches: they not only provide temporal logic model checking, but also bisimulation (resp. simulation) analysis, that is we can establish whether two processes have equivalent behaviors (resp. whether one of the two includes the behavior of the other).

In Figure 1 we present a framework, for the design and verification of WSs using process algebras [6] (*e.g.* CCS, $\pi$-calculus, LOTOS); in this paper we focus on LOTOS, one of the most expressive process algebra. We provide a two-way mapping between BPEL/WSDL and LOTOS, and general guidelines for translations between BPEL/WSDL and a process algebra. Respect to the quoted previous works, we study also the direction from a formal language to BPEL. Using the two-way mapping, that allows an automatic translation between the two languages, two choices are available: designing in BPEL and verifying with a process algebra, designing and verifying in a process algebra. These two approaches are not alternative, but they can be combined in the same development.

**Designing in BPEL and verifying with a process algebra.** Going from BPEL to a PA allows us the verification step in PA, and the converse allows to see the counterexamples directly in BPEL, hopefully even in the visual interface for designing BPEL services. Obviously one can correct in PA, and the BPEL corrected code is automatically generated. This approach is useful also for reverse engineering issues, and when we want to verify BPEL services developed by others.

**Designing and verifying in a process algebra.** We point out that using the mapping we can automatically obtain BPEL/WSDL specifications. To our knowledge this is the first work in this direction. As advocated in a previous work [26], being simple, abstract and formally defined, PAs make it easier to specify the message exchange between WSs, and to reason on the specified systems. They are especially worthy as a first description step because they enable one to analyze the problem at hand, to clarify some points, to sketch a (first) solution using an abstract language (then dealing only with essential concerns), to have at one disposal a formal description of one or more services-to-be, therefore adequate to use existing reasoning tools to verify and ensure some temporal properties (safety, liveness and fairness properties), behavior equivalences (bisimulation), and execution traces. Process algebras design allows the distributed development and software reuse.

Because process algebras support simulation and bisimulation analysis, we can apply to WSs a well-know design method, the *hierarchical refinement* [17,16]: intuitively we start with an abstract description of a process and we refine it iteratively, obtaining at each step a less abstract one. At each stage, using simulation and bisimulation we can verify the correspondence between the current version and the previous (more abstract) one. It can be applied also in the BPEL modelling of WSs, using the two-way mapping.
Moreover we argue, with a simple consideration, that the simulation can be part of the problem of automatic composition of services. Another use of bisimulation is to check the redundancy of service in a community.

In Section 3 we focus on the two-way mapping between LOTOS to BPEL and we give the guidelines formalizing the translation between process algebras and BPEL. In Section 4 we illustrate the features provided by our approach: temporal logic model checking, execution traces, simulation, bisimulation. We discuss the hierarchical refinement and other problems that we can solve using a process algebra representation for WSs and a bisimulation analysis. In Section 5 presents related works and motivates our contribution with respect to them. We draw up concluding remarks in Section 6 and we mention some future works.

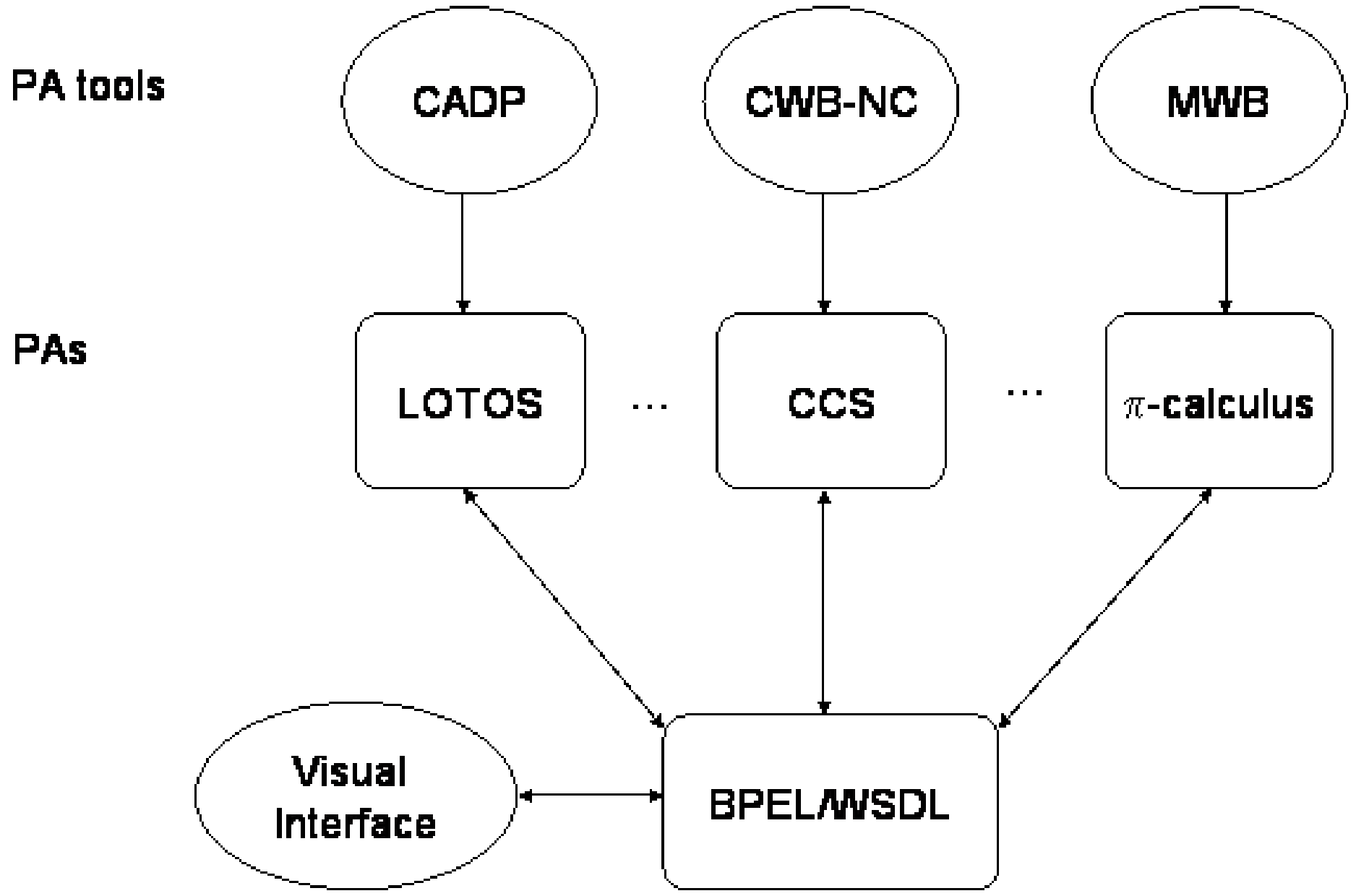

**Fig. 1.** Proposal overview

## 2 Preliminaries

### 2.1 LOTOS in a Nutshell

LOTOS is a specification language for distributed open systems normalized by the ISO [15]. It combines two specification models: one for static aspects (data and operations) which relies on the algebraic specification language ACT ONE [8] and one for dynamic aspects (processes) which draws its inspiration from the CCS [21] and CSP [13] PAs.

**Abstract Datatypes** LOTOS allows the representation of data using algebraic abstract types. In ACT ONE, each *sort* (or datatype) defines a set of *operations* with arity and typing (the whole is called *signature*). A subset of these operations, the *constructors*, are sufficient to create all the elements of the sort. *Terms* are obtained from all the correct operation compositions. *Axioms* are first order logic formulas built on terms with variables; they define the meaning of each operation appearing in the signature.

**Basic LOTOS** This PA authorizes the description of dynamic behaviors evolving in parallel and synchronizing using rendez-vous (all the processes involved in the synchronization should be ready to evolve simultaneously along the same action). A process $P$ denotes a succession of actions which are basic entities representing dynamic evolutions of processes; a process can be recursive. An action in LOTOS is called a *action* (also called event, channel or game in other formalisms). The symbol **stop** denotes an inactive behavior (it could be viewed as the end of a behavior) and the **exit** one depicts a normal termination. The specific **i** action corresponds to an internal (unobservable) evolution.
Now, we present LOTOS behavioral operator. The prefixing operator $G;B$ proposes a rendez-vous on the action $G$, or an independent firing of this action, and then the behavior $B$ is run. The non deterministic choice between two behaviors is represented using []. LOTOS has at its disposal three *parallel composition* operators. The general case is given by the expression $B_1$ $|[G_1, ..., G_n]|$ $B_2$ expressing the parallel execution between behaviors $B1$ and $B2$. It means that $B1$ and $B2$ evolve independently except on the actions $G_1$, ..., $G_n$ on which they evolve at the same time firing the same action (they also synchronize on the termination **exit**). Two other operators are particular cases of the former one to write out interleaving $B_1|||B_2$ which means an independent evolution of composed processes $B_1$ and $B_2$ (empty list of actions), and full synchronization $B_1||B_2$ where composed processes synchronize on all actions (list containing all the actions used in each process). Moreover, the communication model proposes a multi-way synchronization: $n$ processes may participate to the rendez-vous.

The disabling operator $B_1[>B_2$ model the interruption: the behavior $B_1$ could be interrupted at any moment by the behavior $B_2$; when $B_1$ is interrupted, $B_2$ is executed (without having interruptions).

**Full LOTOS** In this part, we describe the extension of basic LOTOS to manage data expressions, especially to allow value passing synchronizations. A process is parameterized by a (optional) list of formal actions $G_{i \in 1..m}$ and a (optional) list of formal parameters $X_{j \in 1..n}$ of type $T_{j \in 1..n}$. The full syntax of a process is the following:

$$\textbf{process}\ P\ [G_0, ..., G_m]\ (X_0{:}T_0, ..., X_n{:}T_n) : func := B\ \textbf{endproc}$$

where $B$ is the behavior of the process $P$ and $func$ corresponds to the functionality of the process: either the process loops endlessly (**noexit**), or it terminates (**exit**) possibly returning results of type $T_{j\in 1..n}$ ($\mathbf{exit}(T_0, ..., T_n)$).

Action identifiers are possibly enhanced with a set of parameters (offers). An *offer* has either the form $G!V$ and corresponds to the emission of a value $V$, or the form $G?X{:}S$ which means the reception of a value of type $S$ in a variable $X$. A single rendez-vous can contain several offers.

A behavior may depend on Boolean conditions. Thereby, it is possible that it be preceded by a guard [*Boolean expression*] $\rightarrow B$. The behavior $B$ is executed only if the condition is true. Similarly, the guard can follow a action accompanied with a set of offers. In this case, it expresses that the synchronization is effective only if the Boolean expression is true (*e.g.*, $G?X$`:Nat[`$X$`>3]`). In the sequential composition operator, the left-hand side process can transmit some values (**exit**) to a process $B$ (**accept**):

$$... \ \mathbf{exit}(X_0, ..., X_n) \gg \mathbf{accept}\ Y_0{:}S_0, ..., Y_n{:}S_n\ \mathbf{in}\ B$$

To end this section, let us say a word about CADP[1], a toolbox that supports developments based on LOTOS specifications. It proposes a wide panel of functionalities from interactive execution to formal verification techniques (minimization, bisimulation, proofs of temporal properties, compositional verification, etc).

### 2.2 Other Process Algebras

Numerous processes algebras have been proposed: CCS [21], CSP [13], ACP [4] are the basic ones. Extensions are $\pi$-calculus [24], Timed CSP [28]. Although syntactically different, all process algebras share a set of basic and dynamic constructs: actions, sequence, parallel composition, synchronizing actions, non deterministic choice, emission, reception, process, local process, recursive process.

### 2.3 Equivalences between processes

Two process are considered equivalent if their behavior is *indistinguishable* from an external observer interacting with them. In the process algebra community several notions of process equivalence have been proposed. More on the topic can be found in [21]. An approach is *trace-based*: two process are equivalent if they show the same execution traces. A process is contained in another one if the set of its execution traces are included in the set of execution traces of the other. Another approach is *tree-based*: two process are equivalent if they have equivalent execution trees, that is they simulate each other (they bisimulate). A process is simulated by another one if all its behaviors are contained in the behaviors of the other. A group of process running concurrently are simulated by another group of process running concurrently if all their behaviors are contained in the behaviors of the other. It is known that simulation implies containment. As example let us discuss Figure 2.

The left process corresponds in basic LOTOS to $a; (b[]c)$, the right one to $a; b\ []\ a; c$. They have the same traces ($ab$ or $ac$), and so they are trace-equivalent. They do not bisimulate each other; after doing $a$ the left process will do either $b$ or $c$, while the right process on doing $a$, it will either choose to move in a state from which it does $b$ or in a state from which it

[1] `http://www.inrialpes.fr/vasy/cadp/`

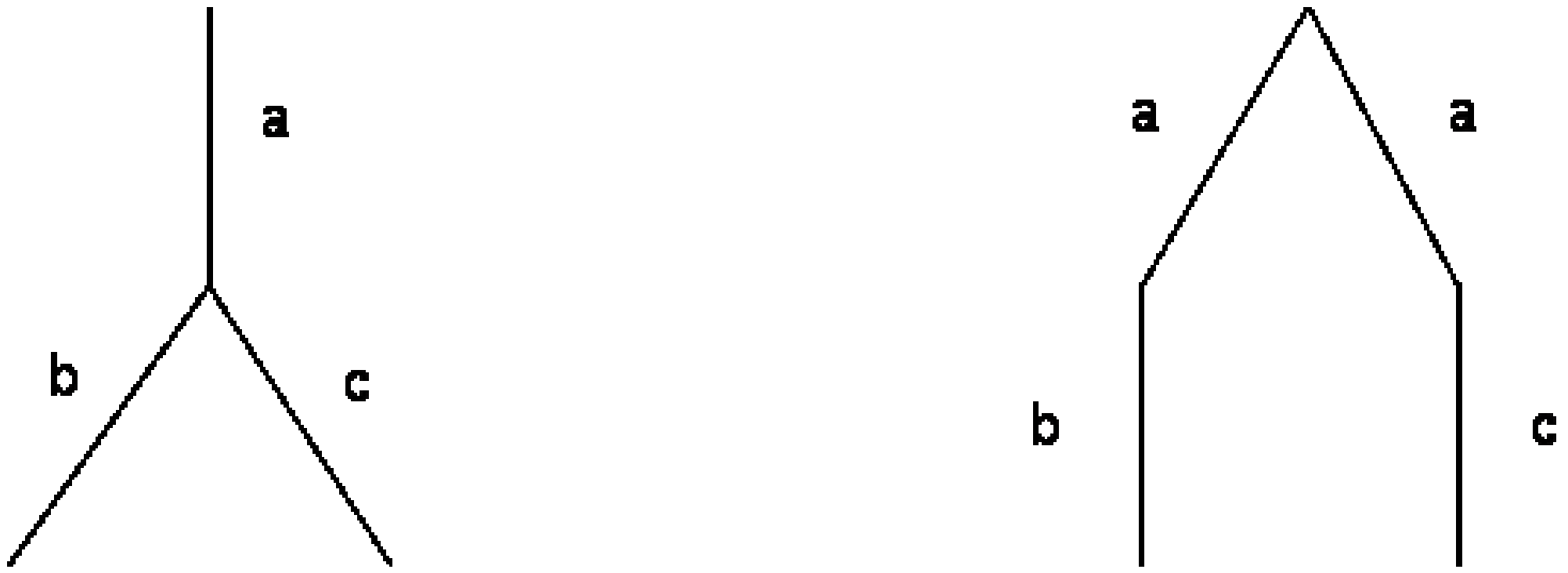

**Fig. 2.** Processes Equivalences; $a$ is a ticket purchase, $b$ ticket use for the match, $c$ a ticket change.

does$c$; depending on this choice, it cannot do one of the two actions whereas the left process leaves both possibilities open. Let $a$ is a ticket purchase, $b$ ticket use for the match, $c$ a ticket change; the left process allows to change the ticket after the purchase, the right one does not.

## 3 The two-way mapping between LOTOS and BPEL

In this section we show the two-way mapping between LOTOS, a process algebra that allows data handling, and BPEL. Our goal is showing a two-way mapping between the two languages, that allows an automated translation. For lack of space, it is not possible to introduce the basics of BPEL, XMLSchema, and XPath. Accordingly, the reader who is not used with them should refer to [3,1,2].
When it is possible, we present together both directions of the mapping. While the translation from BPEL to LOTOS implicitly preserves the BPEL structure, the converse does not: LOTOS allows to use the construct in very flexible manner, BPEL does not. In the LOTOS design we have to be careful, if we want a simple automatic translation, to write behavior structurally similar to BPEL ones. For example in BPEL a service can communicate only with other services, there is no message exchange inside a service. In LOTOS instead, as in all process algebras, there are no constraints about this. In order to obtain a simple automatic translation from a process algebra, we have to follow this simple rule in the design. The details of other similar rules will be given during the explanation. We remark that in our framework, when we design and correct in BPEL, the LOTOS-BPEL direction is free from this problem: we start from LOTOS code, that is BPEL-like structured, because directly obtained by the translation from BPEL.
In our presentation we refer to Table 1, where we show sample code of both languages; the correspondence is about both direction of the mapping. Figure 3 gives a very general picture. We show the mapping of basic construct, dynamic behavior, data definition and handling, and fault, compensation, event handlers. Finally we give general guidelines for translations between PAs and BPEL. An example of translation using this mapping is given in [9].

<table>
<tr><th>Sample BPEL Code</th><th>Sample LOTOS Specification</th></tr>
<tr><td>< ... act1 ... ><br></act1><br><assign ... ><br><copy><br><from expression="5"/><br><to var="x"/><br><copy><br></assign><br>< ... act2 ... ><br></act2></td><td>..act1..; exit(5) $\gg$<br>accept x:Nat in ..act2..</td></tr>
<tr><td><receive ... variable="m"><br></receive</td><td>g?m:Nat;</td></tr>
<tr><td><reply ... variable="m"><br></reply></td><td>g!m:Nat;</td></tr>
<tr><td><invoke ... invar="mS"<br>outvar="mR"><br></invoke></td><td>gS!mS:Nat; gR?mR:Nat;</td></tr>
<tr><td><pick ... ><br><onMessage ... variable="m1"><br>< ... act1 ...><br></onMessage><br><onMessage ... variable="m2"><br>< ... act2 ... ><br></onMessage><br></pick></td><td>(g1?m1:Nat; ..act1..) []<br>(g2?m2:Nat; ..act2..)</td></tr>
<tr><td><sequence ...><br>< ... act1 ... ><br>< ... act2 ... ><br></sequence></td><td>..act1..; ..act2..</td></tr>
<tr><td><flow ... ><br>< ... act1 ... ><br><source linkname="link1"<br>condition="cond1"/><br></act1><br>< ... act2 ... ><br><target linkname="link1"/><br></act2><br></flow></td><td>..act1..;<br>([cond1]->link1 !1; []<br>[not(cond1)]->link1 !0;)<br><br>||<br><br>( link1 ?x:Bool;<br>([x=1]->..act2.. [] [x=0]->i;) )</td></tr>
<tr><td><switch><br><case condition=<br>"bpws:getVariableData(x)>=0"><br><..act1..><br></..act1..><br></case><br><otherwise><br><..act2..><br></..act2..><br></otherwise><br></switch></td><td>[x>=0] -> ..act1..;<br>[]<br>[x<0] -> ..act2.. ;</td></tr>
<tr><td><while condition=<br>"bpws:getVariableData(x)>=0"><br><..act1..><br></..act1..><br></while></td><td>proc while1 [..](..) :=<br>[x<0]-> i;<br>[]<br>[x>=0]->..act1..; while1[..](..)<br>endproc</td></tr>
</table>

**Table 1.** The BPEL-LOTOS two-way mapping examples

### 3.1 General Outline

An external view of interacting WSs shows processes/services running concurrently. Such a kind of global system in LOTOS is described using LOTOS main behavior (that is the outermost process): it instantiates processes composed in parallel and synchronizing on all actions representing their interactions. At the basis of our mapping there is the correspondence between LOTOS actions and BPEL interactions. BPEL services and LOTOS processes instantiated in the main process correspond to each other. The direction from BPEL to LOTOS is straightforward: we simply automatically build a main behavior containing the instantiation of all the processes (each of them correspond to a service), in the manner described above. About the other direction, from LOTOS to BPEL, the LOTOS programmer have to respect this rule: he have to write the main behavior simply instantiating all the processes representing services, in the usual manner.

To describe behaviors, in LOTOS we have the process definition, in BPEL the service description. In LOTOS a defined process can be instantiated (with action passing, that renames the name of action in the definition, and parameter passing). From LOTOS to BPEL, we use the behaviors specified in the process definition to generate the BPEL service description with the names of partner links, port type, operations, variables. From BPEL to LOTOS, we use the service description to generate, including the names of actions, both the process definition and the process instantiations. we have a process instantiation if the process represents a scope or a while.

We do not consider bindings issues. For the data type definitions in BPEL/WSDL we have XMLSchema, in LOTOS we can define abstract data types. In LOTOS we initialize the data structures defined with the type construct at the beginning of the main process.

To summarize the main process first initializes data, then instantiates the process/services running concurrently.

### 3.2 LOTOS actions and BPEL Basic Activities/Interactions

At the core of BPEL process model is the notion of peer-to-peer interaction between partners described in WSDL. All BPEL basic activities perform interactions between WSs. An interaction is characterized by the partner link, the port type, and the operation involved in the two communicating partners (each partner defines these three elements for each interaction). In parallel, LOTOS has at its disposal the notion of action to represent dynamic evolutions and of rendez-vous to describe synchronizations among processes. Consequently, and roughly speaking, when process/services are instantiated LOTOS synchronizing actions are equivalent to BPEL interactions. When the process representing a service is defined, an action is simply an emission or to an reception. The name of the action stores information (parter link, port type, operation in BPEL, process and action names in LOTOS) on the receiver in the emission case, on the sender in the reception case. This name can contain a description of the interaction (e.g. request, notification, cancellation). When we instantiate, we have to compose the names of the action of both interacting processes/services; we consider two synchronizing action, we concatenate their definition name, and we give the concatenated name to both.

Let us go forward in more details. Starting the mapping from BPEL, in order to build the name of LOTOS action, we use the information in *partner link*, *port type*, *operation* attributes in the *receive*, *reply*, and *invoke*. Let a partner 1 (resp. 2) have a partner link $pl_1$ (resp. $pl_2$), a port type $p_1$ (resp. $p_2$), an operation $o_1$ (resp. $o_2$), and a variable $v_1$(resp. $v_2$) associated with the exchanged message. Let a reservation request the object of the partner 1 message, and a availability response the object of the partner 2 message. Then the process associated with the partner 1 has *in the definition* the action $pl_1_p_1_o_1_resReq$ and the process for the partner 2 the action $_pl_2_p_2_o_2_avResp$. When the two process are instantiated in the main behavior, the name of their synchronized action is $pl_1_p_1_o_1_resReq_pl_2_p_2_o_2_avResp$, and $v_1$ (resp. $v_2$) is the parameter of the action for the partner 1 (resp. 2). Moreover if we have a *message* with $N$ *part* tag, in LOTOS we have an action with $N$ parameters, one for each part of the message.
Starting from LOTOS instead, we extract the port type, operation and message definitions analyzing the names of LOTOS actions in the instantiated processes. For example if we have two actions $pl_1_p_1_o_1_pl_2_p_2_o_2$ and $pl_1_p_1_o'_1_pl_2_p_2_o_2$, we conclude that we have the service 1 with partner link $pl_1$, port type $p_1$ and operations $o_1$ and $o'_1$, and a service 2 with partner link $pl_2$, port type $p_1$ and operation $o_2$. Moreover we know thar there are two interactions: one between service 1 in partner link $pl_1$, port type $p_1$, operation $o_1$, and service 2 in partner link $pl_2$, port type $p_2$, operation $o_2$; the other interaction is between service 1 in partner link $pl_1$, port type $p_1$, operation $o'_1$, and service 2 in partner link $pl_2$, port type $p_2$, operation $o_2$.

The reception of a message is expressed using the *receive* activity in BPEL and using a action with a reception in all its parameters in LOTOS.

In BPEL, the emission is written with the *reply* or the *asynchronous invoke* activity whereas in LOTOS we use a action with an emission in all its parameters. The BPEL *synchronous invoke*, performing two interactions (sending a request and receiving a response) corresponds in LOTOS to an emission followed immediately by a reception. In LOTOS we have two different actions, because we have two interactions in BPEL; the names of actions share the same partner link, the same port type, the same operation but they differs only by a letter $S$ or $R$ at the end (representing the emission and the reception of the invoke). Using this rule we can distinguish in the LOTOS code when a contiguous emission-reception is an invoke.

### 3.3 Structured Behaviors

Now we introduce the mapping for LOTOS dynamic constructs and BPEL structured activities.
The *pick* BPEL activity is executed when it receives one message defined in one of its *onMessage* tag or when it is fired by an *onAlarm* event; we cannot model the latter case because basic LOTOS does not have the notion of time. The equivalent construct in LOTOS is obtained using the non deterministic choice, in which the first action of each branch is a reception; it is chosen the branch whose beginnig reception is performed first. In the LOTOS modelling, if we use the non deterministic choice with an emission as first action, the an automatic translation to BPEL become very difficult. For example the following LOTOS behavior, because the $a$ is an emission, does not have a straightforward translation in BPEL:

```
... a!x:Nat; b?x:Nat; [] c?x:Nat; b?x:Nat;..
```

When we design in a process algebra, we have to think to BPEL code structure, in order to simplify the automatic translation.

The *sequence* activity in BPEL match with the LOTOS prefixing operator ';'.

In BPEL we have the *flow* activity, in LOTOS the full synchronization constructs '||'. Because in BPEL we cannot have interaction inside a service, therefore we do not have synchronizations in a parallel composition inside a process representing a service.

The mapping about the *link* tag is more complicated, because LOTOS does not have an explicit construct of dependence relation between concurrent actions. In BPEL we specify with the *source* tag the activity that has to occur first, and with the *target* tag the dependent activity. In LOTOS we have an action for each link. These actions are put after the end of the source behavior, and before the beginning of the target one; the two behaviors synchronizes on these actions, that is they have to execute them at the same time. In this way we are sure that the source behavior is completed before the beginning of the target one. In Table 1, in the flow sample, activity *act2* can be executed only both after executing activity *act1* and the condition *cond1* is true; in the BPEL code, this condition is specified by the *transitionCondition* attribute. In LOTOS after executing $act1$, we execute the action $link1$, representing the link, and assign to its parameter the value 1 if the condition *cond1* is true, 0 otherwise; $act2$ can be executed only if the condition is true and only after $act1$, because it can be executed only after the action $link1$. If we cannot execute $act2$, we have to choices: just to skip $act2$ (in LOTOS we do nothing putting an $i$ action, in BPEL we set $suppressJoinFailure = yes$), or to thow a fault (in LOTOS we put a $fault$ action, in BPEL we set $suppressJoinFailure = no$). More on the links can be found in the Section 3.7.

The *switch* tag defines an ordered list of *case* tag. A case corresponds to a possible activity which may be executed. The condition of a case is a Boolean expression on variables. In our process algebra we have a standard pattern combining guarded expression and non deterministic choice, very often used in the design with LOTOS.

To define an environment with own local variables and with own handler of faults and events, in BPEL there is the *scope* activity, in LOTOS the concept of local process. The process corresponding to the scope is local to the process representing the outer scope. The outermost scope in BPEL is the global one. We deal with this activity in Section 3.5

The *while* BPEL tag and LOTOS recursive processes correspond to each other. The condition of the *while* is the exit condition of the recursive process. The behavior of this recursive process matches exactly the body of the BPEL loop, and conversely. The recursive process is instantiated by the process corresponding to the scope that contains the while.

### 3.4 Data Descriptions

In this subsection, we are going to discuss three levels of data representation in LOTOS and BPEL: data type definitions, XPath and LOTOS, data manipulation.

**Data type definitions** LOTOS allows to define abstract data types, that is data domains and operations on them (e.g. a list with operations: add an element, extract the first one etc.); many basic types (char, natural, etc.) are already defined. In BPEL, types are described using XMLSchema; elements can be simple (lots are already defined) or complex (composed by other elements). A simple element and LOTOS basic data type corresponds each other; moreover we can use the *rename* construct in LOTOS, for example to rename the type string with 'lastname'. We have a complex element in XMLSchema and abstract data type in LOTOS, having one data type for each element composing the complex one.
In XMLSchema complex elements can be composed in different manners, depending from the indicators that establish the order, the number of occurrences of simple elements.
**Order Indicators**. The indicator *all*: each element occurs exactly once, in any order. In LOTOS we can define the abstract data type *list*. The element of the list, that can be added in any order, are the element in the complex type. The indicator *choice*: an element in a set is chosen. In LOTOS we can define the abstract data type *set*. The element of the set are the element in the complex type. The indicator *sequence*: it specifies the elements and the order in which they have to appear. In LOTOS we can use a list whose elements can be added only in a fixed order, depending on the type.
**Occurrence Indicators**. They are use to define how often an element can occur, in details *maxOccurs* the maximum number of times and *minOccurs* the minimum. In LOTOS we have the constraints on the list with a fixed order.
**Group Indicators**. They define a set of elements, with indicators, that can be referenced in another element. In LOTOS we can simply use the abstract data type of the group in the abstract data type of the element that uses the group. For example, if a 'choice' is referenced in a 'all' indicator, an element of the list is a set.

**Variable declaration and manipulation** In LOTOS, variables are either parameters of processes or parameters of a action. In BPEL, variables can represent both data and messages. They are defined using the *variable* tag (global when defined before the activity part) and their scope may be restricted (local declarations) using a *scope* tag. In LOTOS, only process parameters need to be declared (not necessary for action variables) whereas in BPEL either global and local variables involved in interactions have to be declared. In LOTOS, in local process we can declare local variables.
A BPEL *message* corresponds to a set of action parameters in LOTOS. In particular a BPEL *part* corresponds to a parameter of a action in LOTOS.

The BPEL *assign* tag has three equivalents in LOTOS depending on their use: (i) *let* $X_i$:$T_i$=$V_i$ *in* $B$ means the initialization of variables $X_i$ of types $T_i$ with values $V_i$ ($\forall i \in 1..n$) in the behavior $B$, (ii) $B_1$*; exit(*$Y_i$*)* $\gg$ *accept* $X_i$:$T_i$ *in* $B_2$ denotes the modification of variables $X_i$ (replaced by new values $Y_i$), (iii) $P(X_i)$ is an instantiation of a process or a recursive call meaning assignments of values $X_i$ to the parameters of the process $P$. Conversely, these LOTOS constructs can be mapped into BPEL using *assign*, and more precisely the *copy* tag.

**LOTOS and XPath** In BPEL/WSDL we can define either message or data variables, whose type are XMLSchema data structures (element or complex element). XPath is used in BPEL to manipulate data structures: to select element in a complex one, to get value from a variable, to perform operations (e.g. sum, multiplication). LOTOS data structures are abstract data type that are endowed by operations. For example lists have operations for adding an element, extracting the first element and so on. Natural numbers has sum, multiplication and so on. We can use these operations to manipulate data structures. In BPEL we use XPath as expression language; for example we can query data from a variable, and if the variable is a complex type (e.g. a record), we can select the part of interest and retrieve the value. LOTOS instead is similar to the common programming languages like C: when we write a variable, we have its value directly, as in the assign example of Table 1.

### 3.5 BPEL scope and LOTOS pattern of processes

In BPEL the *scope* tag defines a behavior context (local variables, event handlers, fault handlers, compensation handler) for its primary activity. The primary activity describes the normal behavior of the scope. In LOTOS we can define a pattern of processes that behaves in the same way. We point out that a LOTOS user, in the design of a scope with handlers have to respect this pattern of processes, in order to obtain automatically BPEL code. Vice versa, from BPEL specification we can get the LOTOS one, automatically filling this pattern of processes.

In BPEL we can have nested scope. The outermost scope is the global service. In LOTOS we have the concept of local process. In LOTOS the process/scope is local to the outer process/scope. Each process/scope instantiate the following processes:

*primary activity*: a process *primaryActivity* for the primary activity of the *scope*. In the case of normal termination, its last action is an *end* (to end fault and event handlers); we explain it below.

*event handler*: a process $eventHandlers$, executed in full synchronization with *primaryActivity*, because in BPEL event handlers are concurrent with the primary activity of the *scope* to which the event handler is attached.

*fault handlers*: a process $faultManager$ that catches a fault storing its name, launches the process $Kill$ to terminate the *primaryActivity* and $eventHandlers$, then, depending on the fault name, calls the corresponding process to perform the fault activities.

*compensation handler*: a process for the compensation handler. In BPEL we can have at most one compensation handler in a *scope*. The name of the compensation handler is the name given in the *scope* attribute of the activity *compensate*. This process models the activity of the corresponding compensation handler.

Each process/scope has the following structure (LOTOS pseudo-code):

```
proc scopeName [..](..) :=
 ( (primaryActivity[..](..) || eventHandlers[..](..))
    [> Kill[]()
 )
 |[faultName,end]| faultManager[..](..)
endproc
```

The $eventHandlers$ is concurrent with *primaryActivity*; they both can be interrupted by the process $Kill$, launched by the $faultManager$ when a fault occurs. The process for the compensation handler is called inside a process representing a fault or another compensation handler: in BPEL it can be invoked, using the *compensate* tag, only either in a fault handler or in another compensation handler.
Now we introduce the translation about the handlers in details.

**Fault Handlers** When a fault occurs in a BPEL scope, all activities in the primary activity and in the event handlers of the scope begin to terminate. Let the name of the fault $faultName$. In LOTOS we define a process *faultManager* running concurrently respect the process representing the scope, synchronizing on the actions *fault* and *end*. The *fault* action has parameter to communicate the name of the fault; the *end* does not have parameters because we do not need to send or to receive messages, but only to communicate an event. It follows the faultManager definition:

```
proc faultManager [fault, end] (faultName:String) :=
```

```
  ( fault?faultName:String;  Kill;
    [faultName1]-> faultProc1[..](..)
    [faultName2]-> faultProc2[..](..)
    ..
  )
  [] end;
endproc
```

If the scope terminates without faults, the process representing the scope perform as last action the action *end*, allowing to $faultManager$ to terminate without doing nothing. A fault in BPEL is launched through the tag *throw* (that has with attribute the name of the fault) or as response to an invoke activity; in LOTOS through action $fault$. After this the process $Kill$ is instantiated. This process doing nothing, but terminate *primaryActivity* using the disabling construct '[>'. Finally the process corresponding to the fault name is chosen: for example $faultProc1$ corresponds to the fault $faultName1$.
We consider now the problem of fault propagation and handling. In BPEL, when a fault occurs in a scope $S$ that cannot handle it, $S$ terminates abnormally and the fault is propagated to the next scope up. If $S$ can handle the fault, it terminates normally after executing the fault handler activities. From BPEL to LOTOS translation we know which fault handler will catch a fault by parsing the BPEL files. Similarly, from LOTOS to BPEL translation, by parsing the LOTOS specification we know the fault handler that will catch the fault; if the fault is not caught, we have a *stop* action instead of the *fault* one.

**Compensation Handlers** While a business process is running, it might be necessary to undo one of the steps that have already been successfully completed. To each scope we can optionally associate its compensation handler that undoes the primary activity of the scope; once a scope completes successfully, its compensation handler become ready to run. This can happen in either of two cases: explicit or implicit compensation. We map the compensation handler into a LOTOS process local to the process representing the scope.

**explicit compensation:** It occurs upon the execution of a *compensate* activity, that can occur inside a fault handler or a compensation handler of the scope immediately enclosing the scope to be compensated; the *compensate* activity has an attribute *scope* whose value specifies the name of the scope to be compensated. The *compensate* activity is modelled in LOTOS by a call to the process representing the compensation handler associated with the scope.

**implicit compensation:** It occurs when there is a fault handling. Let $A$ be a scope, and $B$ an its nested compensatable scope. Consider the following scenario: $B$ is completed successfully, but an another activity in $A$ throws a fault. Implicit compensation ensures that whatever happened in scope $B$ get undone by running its compensation handler. Therefore, the implicit compensation of a scope goes through all its nested scopes and runs their compensation handlers in reverse order of completion of those scopes. We can map this mechanism in LOTOS by calling in the same order the processes representing the compensation handlers; all this calls are executed in $faultManager$ of $A$ before the beginning of the fault activities. Obviously we have to store the order in which the scopes are completed. We can use a queue data structure in LOTOS to do it; this queue has global visibility and it is updated when a scope completes or if a scope is compensated. The process $faultManager$ of $A$ can use this structure to know the order of completion.

**Event Handlers** It is modeled in LOTOS by a recursive process, concurrent to the primary activity of the scope in which is contained; in this way we can map the BPEL semantics of the event handler: it can accept messages an arbitrary number of times, until the scope ends. We cannot model the *onAlarm* tag, because in LOTOS there is no notion of time. It follows the structure of *eventHandlers*:

```
proc eventHandlers [onMessage1, onMessage2,..] (..) :=
  ( ( fault?faultName:String;
    (onMessage1?m1:T; ..act_m1..;) []
    (onMessage2?m2:T; ..act_m2..;) []
    ..
    )
    eventHandlers [onMessage1, onMessage2,..] (..);
  )
  [] end;
endproc
```

The action *onMessage1* represents the reception of a message $m1$, whose type is $T$. After receiving the message, the corresponding activity *act_m1* is executed. Then the process recursively calls itself, and it ends when an *end* interaction happens.

### 3.6 Guidelines for translation between a PA and BPEL

Slightly modifying the mapping for LOTOS, we easily obtain a mapping for other process algebras. In fact, while syntactically different, they share many concepts: the emission (message sending), the reception (message receiving), the sequence of actions, the concurrency of actions (parallel composition) and their synchronization, the processes and local ones, non deterministic choice of actions. In Figure 3 we give the outline of the correspondences. We

| **BPEL concept** | **Process Algebra concept** |
|---|---|
| service (process) | process |
| scope | local process |
| interaction | synchronizing action |
| receive | reception |
| reply | emission |
| asynchronous invoke | emission |
| synchronous invoke | emission immediately followed by a reception |
| sequence | sequence construct |
| flow | parallel composition |
| while | recursive process |
| pick | non deterministic choice |

**Fig. 3.** The BPEL-PA correspondences

remark that for modelling in a PA, if one wants a simple automatic translation, the PAs processes have to respect the BPEL structure, as in LOTOS.
If the PA does not support the data definition and handling, the mapping is slightly different:

in this case the messages are tokes, and we cannot distinguish between parts in a message. In details from BPEL to PAs we use the information in *partner link*, *port type*, *operation* attributes in order to build the name of a action. If a partner 1 (resp. 2) has a partner link $pl_1$ (resp. $pl_2$), a port type $p_1$ (resp. $p_2$), an operation $o_1$ (resp. $o_2$), and a variable $v_1$(resp. $v_2$) associated with the exchanged message, then the name of the action in the instantiation is $pl_1_p_1_o_1_v_1_pl_2_p_2_o_2_v_2$. In another words, now the message it is not a parameter of the action, but it is a part of the action name: it characterizes the interaction. It is worth noting that for the translation from PAs to BPEL, if the designer respects such a structure, partner links, port types, and operations involved in the BPEL interactions can be deduced automatically from PAs actions. Otherwise, the user have to give the names manually.

### 3.7 An Example

Now we give an example of translation using the "Loan Approval Process", taken from [3]. The service interacts with the services of customer, loan assessor, loan approver. A loan amount is proposed by the customer. If the amount is lower than $10000, and if the loan assessor gives a "low-risk" assessment, the loan is approved; otherwise the approver have to make the decision. In any case the service communicates to the customer the decision.

If follows the WSDL message definition:

```
<message name="creditInformationMessage">
   <part name="firstName" type="xsd:string"/>
   <part name="name" type="xsd:string"/>
   <part name="amount" type="xsd:integer"/>
</message>

<message name="approvalMessage">
   <part name="accept" type="xsd:string"/>
</message>

<message name="riskAssessmentMessage">
   <part name="level" type="xsd:string"/>
</message>

<message name="errorMessage">
   <part name="errorCode" type="xsd:integer"/>
</message>
```

In the business process defined below, the interaction with the customer is represented by the initial `<receive>` and the matching `<reply>` activities. The use of risk assessment and loan approval services is represented by `<invoke>` elements. All these activities are contained within a `<flow>`, and their (potentially concurrent) behavior is staged according to the dependencies expressed by corresponding `<link>` elements. Note that the transition conditions attached to the `<source>` elements of the links determine which links get activated. Because the operations invoked can return a fault of type "loanProcessFault", a fault handler is provided. When a fault occurs, control is transferred to the fault handler, where a `<reply>` element is used to return a fault response of type "unableToHandleRequest" to the loan requester. We omit the details in the code not involved in the translation.

```
<process name="loanApprovalProcess"
         suppressJoinFailure="yes" ...>
 ...
<!-- variables declaration>
<variables>
 <variable name="request"
    messageType="lns:creditInformationMessage"/>
 <variable name="risk"
    messageType="lns:riskAssessmentMessage"/>
 <variable name="approval"
    messageType="lns:approvalMessage"/>
 <variable name="error"
    messageType="lns:errorMessage"/>
</variables>

<!-- fault handler definition>
<faultHandlers>
 <catch faultName="lns:loanProcessFault"
        faultVariable="error">
    <reply partnerLink="customer"
           portType="lns:loanServicePT"
           operation="request"
           variable="error"
           faultName="unableToHandleRequest"/>
 </catch>
</faultHandlers>

<!-- behavior definition>
<flow>
  <links>
     <link name="receive-to-assess"/>
     <link name="receive-to-approval"/>
     <link name="approval-to-reply"/>
     <link name="assess-to-setMessage"/>
     <link name="setMessage-to-reply"/>
     <link name="assess-to-approval"/>
  </links>
  <receive partnerLink="customer"
           portType="lns:loanServicePT"
           operation="request"
           variable="request" createInstance="yes">
     <source linkName="receive-to-assess"
       transitionCondition=
       "bpws:getVariableData('request','amount')
                                   < 10000"/>
     <source linkName="receive-to-approval"
       transitionCondition=
```

```
      "bpws:getVariableData('request','amount')
                              >=10000"/>
  </receive>
  <invoke  partnerLink="assessor"
           portType="lns:riskAssessmentPT"
           operation="check"
           inputVariable="request"
           outputVariable="risk">
     <target linkName="receive-to-assess"/>
     <source linkName="assess-to-setMessage"
       transitionCondition=
       "bpws:getVariableData('risk','level')='low'"/>
     <source linkName="assess-to-approval"
       transitionCondition=
       "bpws:getVariableData('risk','level')!='low'"/>
  </invoke>

  <assign>
     <target linkName="assess-to-setMessage"/>
     <source linkName="setMessage-to-reply"/>
     <copy>
        <from expression="'yes'"/>
        <to variable="approval" part="accept"/>
     </copy>
  </assign>

  <invoke  partnerLink="approver"
           portType="lns:loanApprovalPT"
           operation="approve"
           inputVariable="request"
           outputVariable="approval">
     <target linkName="receive-to-approval"/>
     <target linkName="assess-to-approval"/>
     <source linkName="approval-to-reply" />
  </invoke>
  <reply  partnerLink="customer"
          portType="lns:loanServicePT"
          operation="request"
          variable="approval">
    <target linkName="setMessage-to-reply"/>
    <target linkName="approval-to-reply"/>
  </reply>
</flow>

</process>
```

Now we show the LOTOS code. Due to the lenght of the action names, we shorten them the following way:

```
request <- customer_lns:loanServicePT_request
check_S <- assessor_lns:riskAssessmentPT_check_S
check_R <- assessor_lns:riskAssessmentPT_check_R
approve_S <-approver_lns:loanApprovalPT_approve_S
approve_R <-approver_lns:loanApprovalPT_approve_R
```

To specify parts of the message in BPEL we have the *part* tag, in LOTOS the action parameters. Each parameter of the actions models a part of the exchanged message. For example the BPEL variable `request` has three parts: `firstName` (string), `name` (string), `amount` (integer). In LOTOS we have three parameters with same corresponding names and types.

The *Approval* process definition, without considering the dependencies between activities expressed by links:

```
proc Approval [request, check_S, check_R, approve_S, approve_R]
              (firstName:String, name:String, amount:Integer,
               level:String, accept:String) :=
 \\ receive
 request ?firstName:String ?name:String ?amount:Integer

 ||

 \\invoke
 check_S !firstName:String !name:String !amount:Integer;
 check_R ?level:String

 ||

 \\ assign
 exit(yes) accept accept:String in {

   \\ invoke
   approve_S !firstName:String !name:String !amount:Integer;
   approve_R ?accept:String
   ||

   \\ reply
   request !accept:String
 }
endproc
```

It follows the *Approval* process definition, now considering the dependencies between activities expressed by links:

```
proc Approval [request, check_S, check_R, approve_S, approve_R,
               link_receive-to-assess, link_receive-to-approval,
```

```
              link_assess-to-setMessage, link_assess-to-approval,
              link_setMessage-to-reply, link_approval-to-reply]
             (firstName:String, name:String, amount:Integer,
              level:String, accept:String,
              b1:Bool, b2:Bool, b3:Bool, b4:Bool) :=

\\receive
request ?firstName:String ?name:String ?amount:Integer;
( [amount < 10000] -> link_receive-to-assess;
                      \\ source receive-to-assess
  []
  [amount >= 10000] -> link_receive-to-approval!1;
                       \\ source receive-to-approval
)

||

\\ invoke
link_receive-to-assess;   \\ target receive-to-assess
check_S !firstName:String !name:String !amount:Integer;
check_R ?level:String;
( [level = 'low'] -> link_assess-to-setMessage;
                      \\ source assess-to-setMessage
  []
  [level != 'low'] -> link_assess-to-approval!1;
                      \\ source assess-to-approval
)

||

\\ assign
link_assess-to-setMessage;      \\ target assess-to-setMessage
exit(yes) accept accept:String in {
  link_setMessage-to-reply!1;   \\ source setMessage-to-reply

  \\ invoke
  \\ target receive-to-approval and assess-to-approval
  (link_receive-to-approval ?b1:Bool   []   link_assess-to-approval ?b2:Bool);
  (link_receive-to-approval ?b1:Bool   []   link_assess-to-approval ?b2:Bool);

  ( [b1=1 AND b2=1]-> ( approve_S !firstName:String !name:String !amount:Integer;
                        approve_R ?accept:String );
   []
    [not(b1=1 AND b1=1)]-> i
  )
  link_approval-to-reply!1;   \\ source approval-to-reply
```

```
   ||

   \\ reply
   \\ target setMessage-to-reply and target approval-to-reply
   (link_setMessage-to-reply ?b3:Bool   []   link_approval-to-reply ?b4:Bool);
    []
   (link_setMessage-to-reply ?b3:Bool   []   link_approval-to-reply ?b4:Bool);
   ( [b3=1 AND b4=1]-> request !accept:String
    []
     [not(b3=1 AND b4=1)]-> i
   )
 }
endproc
```

The case of an activity that depends from two or more other activities is more complicated: the source activities can happen in any sequence. In LOTOS we model this situation using the non deterministic choice in which each branch is a link activity; if an activity depends from $N$ source activities, we repeat the non deterministic choice of all link action $N$ times. In this way the we have to receive $N$ values; we ensure that we receive $N$ different aknowledgement of a source activity execution by the guarded expression before performing the target activity. For example the last `request` has to be executed after `link_setMessage-to-reply` and `link_approval-to-reply` in any order; using the non deterministic choice we ensure that we receive two aknowledgement of execution of the source activity. We ensure that both activities are performed by executing the target activity `request` only if we have $b3 = 1$ (the action `link_setMessage-to-reply` is executed) and $b4 = 1$ (the action `link_approval-to-reply` is executed). If $b3 = 0$ or $b4 = 0$ we simply do nothing, according to the $suppressJoinFailure =' yes'$ attribute in the BPEL process tag. We remark that $b3$ and $b4$ are initialized with 0 by the main process; all LOTOS parameters related with the link actions have to be initialized with 0 (we always follow this rule in the mapping about the links).

The *faultManager* process definition:

```
proc faultManager [fault, end] (faultName:String) :=
 ( fault?faultName:String; Kill;
   [lns:loanProcessFault]-> faultProc1[request](error)
 )
 [] end;
endproc
```

The *faultProc1* process definition:

```
proc faultProc1 [request](error:Integer) :=
  request !errorCode:Integer
endproc
```

Finally we give the LOTOS main process, considering also the links; we concatenate the names of two actions in process definition that constitute an interaction (e.g. the action `request_c` is the interaction composed by the action 'request' (a reception) in the process definition `Approval` and the action 'c' (an emission) in the process definition `Customer`). The two actions are instantiated with the same name. For example the `request` in the process definition `Approval` becomes, in the `Approval` instantiation, `request_c`; the action `c` in the process definition `Customer` is replaced by `request_c` in the instantiation. The actions corresponding to the links are not renamed (they do not model an interaction between services); we always follow this rule in the mapping about the links.

```
proc main [request_c, check_S_ass, check_R_ass, approve_S_app, approve_R_app,
           link_receive-to-assess, link_receive-to-approval,
           link_assess-to-setMessage, link_assess-to-approval,
           link_setMessage-to-reply, link_approval-to-reply, ..]
          (firstName, name, amount, level, accept, ..) :=
  ( ( Approval[request_c, check_S_ass, check_R_ass, approve_S_app, approve_R_app,
               link_receive-to-assess, link_receive-to-approval,
               link_assess-to-setMessage, link_assess-to-approval,
               link_setMessage-to-reply, link_approval-to-reply]
              (firstName, name, amount, level, accept, 0, 0, 0, 0)
      || Customer[request_c,..](..)
      || Assessor[check_S_ass, check_R_ass,..](..)
      || Approver[approve_S_app, approve_R_app,..](..)
    )
    [> Kill[]()
  )
  |[fault,end]|
  faultManager[]()
endproc
```

## 4 Design and verification features

In this section we discuss the features that process algebras provide for design and verification, and we sketch some problems, for a future work, that deal with simulation and bisimulation. Examples of WS developed using process algebras can be found in [26], where a sanitary agency is modelled in CCS, and in [27], where a simple e-commerce application is designed using LOTOS.

**Distributed development and reuse**. The modelling from a process algebra supports the distributed development and the software reuse. In fact for developing a service, one need to know only actions and parameters of the other interacting services. Similarly to develop a WS in BPEL, we need to know only the WSDL specification, not the internal behavior, of the other interacting services.

**Verification features**. The following verification facilities are available at an early stage of the Web services deployment:

- temporal logic model checking, in order to prove properties of the service: liveness (something good happens), safety (bad events do not happens), request-response (a request is always satisfied, also for infinite behaviors), and others. We can verify for example mutual exclusion properties (e.g. if the provider can satisfies only one request among multiple concurrent requests, it satisfies the first confirmed request). If the property is not satisfied, a counterexample is returned.

- bisimulation, to check whether the behaviors of two services or two versions of the same service are equivalent; if they are different, it is shown a counterexample.
- simulation, to check whether the behavior of a services is included by the behavior of other interacting services; if it is not, it is shown a counterexample.
- execution traces of the service (manually or random guided), to understand the behavior of the service. In the verification community (and in [23]), often the simulation name is used to denote execution traces analysis; this is no the case of this paper.

In the case of a process algebra allowing the data handling, it is available:

- data type checking, in the case of LOTOS and other process algebras allowing data handling.
- black box testing: for a class of input values, some properties are satisfied.

Respect previous approach [23,10,22,11], one of the main advantage of using process algebra is the availability of the simulation and bisimulation analysis; they supports the hierarchical refinement design method and the redundancy analysis of a community. Moreover we argue, with a simple consideration, that the simulation can be part of the problem of automatic composition of services. In the rest of the section we discuss briefly these issues, considering them for a future work.

**Hierarchical refinement [17,16]**. It is a well-known method for design development. It proceedes top-down: starting with a highly abstract specification, we construct a sequence of behavior descriptions, each of which refers to its predecessors as a specification, and is thus less abstract than the predecessor. At each stage the current implementation is verified to satisfy its specification. The last description in the sequence contains no abstractions, and constitutes the final implementation. The behavioral equivalence between a specification and its implementation is checked by simulation or by a trace-based equivalence. The advantage of using a two-way mapping, rather than only the direction starting from BPEL, is that we can apply hierarchical refinement also in the BPEL modelling of WS.

**Automatic Composition and Redundancy**. The simulation can be part of the problem of automatic composition of services: intuitively, a service is composable from a bundle of other ones, if it can be simulated by them, that is if its behaviors are contained in their behaviors.

When a community of Web services is used to compose a new service (e.g. [5]), it is useful to know which services in the community are redundant: we can calculate it off-line, using bisimulation. On-line, before starting the composition algorithm, we select services avoiding that two or more equivalent services are activated.

We stress that many applications can be developed using our methodology, especially those involving critical concerns (*e.g.* e-commerce) and therefore which need a formal reasoning and development: auction bargaining, on-line sales, banking systems etc.

## 5 Related Works

We are going to introduce three kinds of related works aiming at: i) specifying WSs at an abstract level using formal description techniques and reasoning on them, ii) using jointly abstract descriptions and executable languages (mainly BPEL), iii) developing WSs from abstract specifications.

At this abstract level, lots of proposals originally tended to describe WSs using semi-formal notations, especially workflows [20]. More recently some more formal proposals grounded for most of them on transition system models (LTSs, Mealy automata, Petri nets) have been suggested [14,23,12,5,19]. With regards to the reasoning issue, works have been dedicated to verifying WS description to ensure some properties of systems [11,7,23,10,22]. Summarizing these works, they use model checking to verify some properties of cooperating WSs described using XML-based languages (DAML-S, WSFL, BPEL, WSCI). Accordingly, they abstract their representation and ensure some properties using ad-hoc or well-known tools (*e.g.* SPIN, LTSA). We have a deeper look in the following of this section at proposals focusing on BPEL.

In comparison to these existing works, the strength of our alternative approach (using PA) is to work out all these issues (description, composition, reasoning) at an abstract level, based on the use of expressive (especially compared to the former proposals) description techniques and adequate tools. The compositionality property of process algebra is also very convenient in one area where composition is one of the main concern.

The second bunch of related work [11,25,10,23,29] deals with mappings between abstract and concrete descriptions of WSs. Let us emphasize that in first attempts [26,27], we have already proposed some guidelines to map process algebra and BPEL. Nevertheless, these guidelines (for CCS and LOTOS) were not defined in details and they deal with a subset of BPEL; in this work we include in the mapping also fault, compensation, and event handlers. Two relevant related works are [10,11]. In the first one, the authors proposed a formal approach to model and verify the composition of WSs workflows using the FSP (Finite State Processes) notation and the LTSA tool. Their paper introduces a translation of the main BPEL structured activities (sequence, switch, while, pick and flow) into FSP processes. In the second one, it is presented an approach to analyze BPEL composite web services communicating through asynchronous messages. They use guarded automata as an intermediate language from which different target languages (and tools) can potentially be employed. They especially illustrate with the use of Promela/SPIN as the formal language and the corresponding model checker. Compared to them, our attempt is more general: (i) we show a two-way mapping, useful to develop WSs and also to reason on deployed ones (the latter direction was the single goal of mentioned related works). All other previous works give only a mapping from BPEL to a formal language. (ii) we consider in the mapping also compensation and event handlers, and we deal with fault handlers explicitly. (iii) we can verify not only temporal logic properties, but also behaviors equivalences between services using bisimulation. Using this facilities we can apply the hierarchical refinement design method to WSs, also in the BPEL modelling.

Finally, the recent proposal of Lau and Mylopoulos [18] argue the use of TROPOS as starting point of WS design, but they do not deal with verification, but requirements issues.

## 6 Concluding Remarks and future work

We present a framework, for the design and verification of WSs using process algebras. We illustrate a two-way mapping between a very expressive process algebra, LOTOS, and BPEL. We give also general guidelines for translations between a process algebra and BPEL. Process algebra allows not only temporal logic model checking, but also a simulation and bisimulation analysis; they allow a design method, hierarchical refinement [17,16], that we can apply to WSs. In fact the two-way mapping allow us to design and verify both in process algebra and in BPEL. In Section 4, we sketch how simulation and bisimulation are involved in the automatic composition of services and in the redundancy check of services. In our opinion, these connections deserve to be studied in a future work, together with the generalization of the mapping to other languages and its implementation. In our current mapping, we do not consider dynamic process instantiation and correlation set. Moreover we do not tackle the problem of the dynamic choice of the partner to talk to (our interactions are established before the conversation between partners starts); for this reason we do not consider BPEL endpoint references. It is interesting to extend the mapping in these directions.